\documentclass[10pt,twocolumn,letterpaper]{article}

\usepackage{iccv}
\usepackage{times}
\usepackage{epsfig}
\usepackage{graphicx}
\usepackage{amsmath}
\usepackage{amssymb}

\usepackage{multirow}

\DeclareMathOperator*{\argmin}{argmin}
\DeclareMathOperator*{\argmax}{argmax}

\usepackage[breaklinks=true,bookmarks=false]{hyperref}

\iccvfinalcopy 

\def\iccvPaperID{****} 

\ificcvfinal\fi

\begin{document}

\title{Conditional Coupled Generative Adversarial Networks for Zero-Shot Domain Adaptation}

\author{Jinghua Wang and Jianmin Jiang\\
Research Institute for Future Media Computing, \\College of Computer Science \& Software Engineering, Shenzhen University, Shenzhen, China\\
{\tt\small {wang.jh@szu.edu.cn, jianmin.jiang@szu.edu.cn}}
}

\maketitle
\ificcvfinal\thispagestyle{empty}\fi

\begin{abstract}
Machine learning models trained in one domain perform poorly in the other domains  due to the existence of domain shift. Domain adaptation techniques solve this problem by training transferable models from the label-rich source domain to the label-scarce target domain.  Unfortunately, a majority of the  existing domain adaptation techniques rely on the availability of  target-domain  data, and thus limit  their applications to a small community across few computer vision problems. In this paper, we tackle the challenging zero-shot domain adaptation (ZSDA) problem, where  target-domain data is non-available in the training stage.  For this purpose, we propose conditional coupled generative adversarial networks (CoCoGAN) by extending the coupled generative adversarial networks (CoGAN) into a conditioning model.  Compared with the existing state of the arts, our proposed CoCoGAN is able to capture the joint distribution of dual-domain samples in two different tasks, i.e. the relevant task (RT) and an irrelevant task (IRT).  We  train   CoCoGAN with both source-domain samples in RT and   dual-domain samples in IRT to complete the domain adaptation. While the former provide   high-level concepts of the non-available target-domain data, the latter carry the sharing correlation between the two domains  in RT and IRT.  To train  CoCoGAN in the absence of  target-domain data  for RT, we propose a new supervisory signal, i.e. the alignment between representations across tasks. Extensive experiments carried out demonstrate that our proposed CoCoGAN outperforms existing state of the arts in image classifications.
\end{abstract}

\section{Introduction}

Most machine learning techniques assume that the training and testing data are from the same domain and follow the same distribution.  
In the real world, however, data samples often originate from different domains. 
For example, the image of an object can be captured by either a RGB sensor or a depth sensor. 
Though the data in two domains may share the high-level concepts, they are significantly different from each other due to the existence of domain shift.
As a result, the models learned in one domain perform poorly in the other \cite{Gretton-2009-covariate-shift}.
Domain adaptation aims to overcome this problem by learning transferable knowledge from the source domain to the target domain.

In general, domain adaptation techniques assume that the labels of data samples are shared by the source domain and the target domain \cite{Gabriela-domain-adaptation-survey}. 
Under such an assumption, many different strategies can be made applicable for domain adaptation. 
Motivated by the theoretical analysis \cite{Shai-ben-david-a-theory-of}, some researchers reduce the domain divergence and improve the performance in  target domain either by minimizing the  discrepancy  of representations between domains \cite{Tzeng2015SimultaneousDT,Long-icml2015-LearningTransferableFeatures,Long-2016:UDA:3157096.3157112} or  by adversarial training \cite{Ganin:2016:DTN:2946645.2946704,Sohn2017UnsupervisedDA,Tran2018JointPA,Sankaranarayanan-cvpr-18a-generate-to-adapt,Liu-nips2016-CoupleGAN}.
Self-ensembling techniques are  proposed  to obtain consistent predictions in two different domains \cite{Laine-temporal-iclr2017,French2018selfensembling,Tarvainen-nips2017-NIPS2017_6719}. 
The encoder-decoder frameworks are also reported in the literature for  many domain adaptation tasks \cite{Chen:2012:MDA:3042573.3042781,Ghifary2016DeepRN}.

\begin{figure}
	\includegraphics[width=0.99 \linewidth]{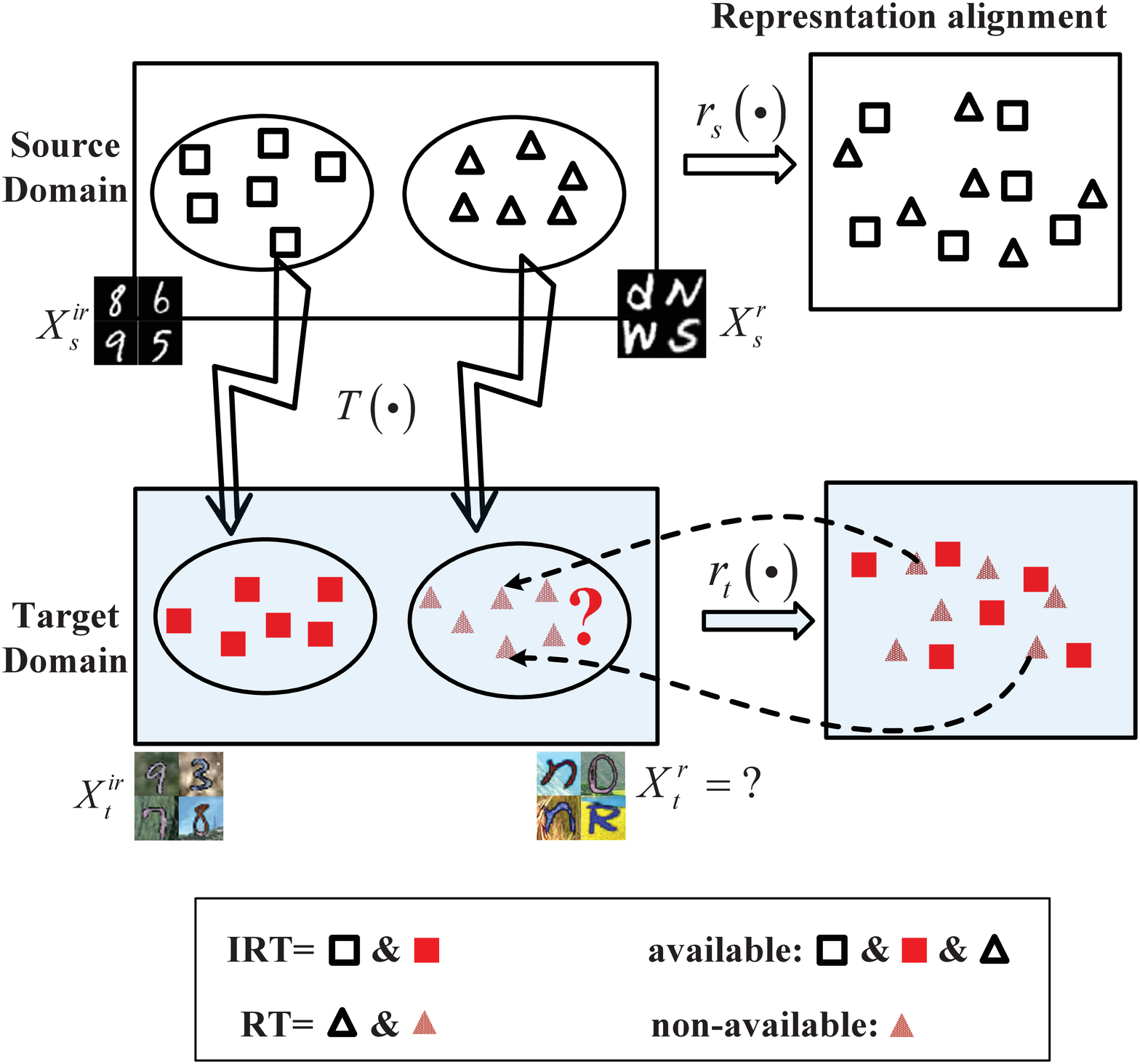}
	\caption{Illustration of an example of ZSDA task, where the RT and IRT are  letter image analysis and digit image analysis, respectively. The source domain consists of gray-scale images and the target domain consists of color images. We first impose the alignment in the source domain based on the available data, then use the alignment in the target domain to guide the training procedure  so that CoCoGAN can search the data space to generate proper target-domain samples of RT.}
	\label{fig:intuitveExample}
\end{figure}

Although the above methods are successful  in various tasks, none of them is applicable in zero-shot domain adaptation (ZSDA) cases where the target-domain data for the task of interest are non-available. 
%
A typical ZSDA example is the personalization procedure of a spam filter before the user start to use the email system, where the target-domain represents a unique distribution of emails received by the user. 
In present, the challenging ZSDA receives increasing attentions over recent years, and the existing methods either learn domain-invariant features \cite{Muandet-Domain-icml2013,Ghifary2015Domain,Li_2018_ECCV,Peng-2018-eccv-zero-shot} or represent the unseen target domain parametrically \cite{Yang-zero-shot-Domain,Kumagai-2018arxiv-ZeroshotDA}.

To achieve more effective ZSDA, we propose a new method to learn target-domain models  not only based on the source-domain samples for the task of interest, but also gain supervision from the dual-domain samples in an irrelevant task.
For simplicity, we denote  the relevant task (task of interest) as RT and the irrelevant task as IRT.
We also denote the source-domain sample set as $ X_s^r $ in RT and $ X_s^{ir} $ in IRT.
Similarly, we use $ X_t^r $  and $ X_t^{ir} $ to represent the target-domain sample set in RT and IRT, respectively.
In this work, we assume that the transformation from the source domain to the target domain is shared by both RT and IRT. Mathematically,  if $ X_t^r=T_r(X_s^r) $ and $ X_t^{ir}=T_{ir}(X_s^{ir}) $, we assume $ T_r=T_{ir} $.
Thus,  while the high-level concepts of the non-available $ X_t^r $ are carried by  $ X_s^r $,   the correlation between two domains can be learned in  IRT  where the dual-domain samples are available. 
Fig. \ref{fig:intuitveExample} illustrates an example of ZSDA, where the source domain consists of gray-scale images and the target domain consists of color images (obtained using the method in \cite{Ganin-icml2015-unsupervised}).  The RT and IRT are  letter image analysis and digit image analysis, respectively.

Our conditional coupled generative adversarial network (CoCoGAN) captures the joint distribution of  dual-domain samples in both RT and IRT  by extending the coupled generative adversarial networks (CoGAN)  \cite{Liu-nips2016-CoupleGAN} into a conditional model with a binary conditioning variable. The proposed CoCoGAN consists of two GANs, i.e. GAN$ _s $ for the source domain and GAN$ _t $ for the target domain. 
The difficulty in training the CoCoGAN for a ZSDA task lies in the non-availability of  $ X_t^r $. Consequently, the GAN$ _t $ tends to be biased towards IRT in the target domain.
We solve this problem by introducing  a new supervisory signal, i.e. the alignment between sample representations across tasks in a given domain. 
Based on the conjugation between the two branches in CoCoGAN, we impose   representation alignment in source domain based on the available data, and expect that the representations of the generated $ X_t^r $ are aligned with   the representations of the available $ X_t^{ir} $   in   target domain, as shown in Fig. \ref{fig:intuitveExample}. 
Specifically, we search the target domain by updating the parameter of GAN$ _t $   to generate a proper non-available  $ X_t^r $ as such that their representations are indistinguishable from the representations of the available  $ X_t^{ir} $.

We highlight our contributions in three-folds. 
Firstly, we propose a new network structure, i.e. CoCoGAN, by  extending the CoGAN into a conditioning model.  The proposed CoCoGAN is able to capture the joint distribution of data samples in two different tasks.
Secondly, we propose a method to train the CoCoGAN for ZSDA tasks by introducing representation alignment across tasks as the supervisory signal. 
Thirdly, in comparison with the work \cite{Peng-2018-eccv-zero-shot},  our new method solves the ZSDA tasks without relying on the correspondences between samples in  the IRT, and thus has more potential applications.

\section{Related Work}
\label{sec:relatedwork}

Domain shift refers to the fact that data samples follow different distributions across domains.
As a result, a solution learned in one domain often performs poorly in another. Domain adaptation techniques solve this problem by learning transferable solutions from a label-rich source domain to a label-scarce target domain, and achieve success in a range of learning tasks, such as image classification \cite{Gebru-fine-grain-iccv2017,Koniusz-cvpr2017,Motiian2017Unified,Wang:2018:VDA:3240508.3240512,Qi-acmmultimedia-2018-aunified} and semantic segmentation \cite{Yang2017Curriculum}.

The most popular strategy for domain adaptation is to eliminate the domain shift and learn invariant features  by either minimizing a well defined criteria or adversarial learning. 
For fine-grained recognition, Gebru et al. \cite{Gebru-fine-grain-iccv2017} minimize a multi-task objective consisting of label prediction loss and attribute consistent loss.
Hu et al. \cite{Hu-deep-metric-cvpr015} learn a deep metric network  to minimize  both the distribution divergence between domains and the marginal Fisher analysis criterion.
Long et al. \cite{Long-icml2015-LearningTransferableFeatures} propose a deep adaptation network to learn domain-invariant features by minimizing the maximum mean discrepancy (MMD) metric.
Tzeng et al. \cite{TzengHZSD14-deepdomainconfusion} introduce a domain confusion loss and an adaptation layer to learn representations which are not only domain invariant but also semantically meaningful. 
Other domain adaptation methods also adopt the MMD metric \cite{Long-2016:UDA:3157096.3157112,Yan-cvpr2017-mind-the-class}. Motivated by the success of generative adversarial networks (GAN) \cite{Goodfellow-gan}, Tzeng et al. \cite{Tzeng2017AdversarialDD} propose a general framework, i.e. adversarial discriminative
domain adaptation (ADDA), to 
combine discriminative modeling, untied weight sharing,
and a GAN loss. In order to transform data from one domain to another in the pixel space, Bousmalis et al. \cite{Bousmalis2017UnsupervisedPD} propose to decouple the processes of domain adaptation  and task-specific classification based on adversarial learning. Yoo et al. \cite{Yoo2016PixelLevelDT} propose an image-conditional  model for transformation from source domain to   target domain in semantic level and apply it to  fashion analysis tasks. 
Shrivastava et al. \cite{Shrivastava2017LearningFS} adopt GAN  for simulated+unsupervised (S+U) learning to improve the realism of the generated images.

All of the above methods rely on the availability of  target-domain data in their training stages, which are factually not always the case in real-world.
For example, we may feel disappointed to see that a computer vision system working well with an existing low-resolution camera under-performs when it is replaced by a high-resolution camera. 
Zero-shot domain adaptation (ZSDA) refers to such a task, where the target-domain data are non-available in the training stage.

Over the past years, many methods are proposed to tackle the ZSDA problem. 
Khosla et al. \cite{Khosla2012Undoing} exploit dataset bias and learn a set of visual world weights which are common to all datasets. Later, Li et al. \cite{Li2017Deeper} use the neural network structure to implement the similar idea. 
Yang and Hospedales \cite{Yang-zero-shot-Domain} predict and describe the unseen target domain by a continuous parametrized vector.
Kodirov et al. \cite{Kodirov-iccv2015-unsupervisedDA-ZSL} solve the domain shift problem with a regularized sparse coding framework.
Kumagai and Iwata \cite{Kumagai-2018arxiv-ZeroshotDA}  introduce the concept of latent domain vectors to characterize different domains and use them to infer the models for unseen domains.

Researchers also propose ZSDA techniques to learn domain-invariant features, which are applicable in not only the source domain but also  the unseen target domain. Muandet et al. \cite{Muandet-Domain-icml2013} propose domain-invariant component analysis (DICA) to  learn an invariant transformation by maximizing the similarity  between different domains.
Ghifary et al. \cite{Ghifary2015Domain} propose multi-task autoencoder (MTAE) to learn transformation from an image to its correspondence in the other domains, and thus obtain domain-invariant features.
Li et al. \cite{Li_2018_ECCV}  propose a conditional invariant adversarial network to minimize the discrepancy of conditional distributions across domains. 
The work that most related with ours is by Peng et al. \cite{Peng-2018-eccv-zero-shot} which learns knowledge from the  dual-domain images in an irrelevant task. However, the work \cite{Peng-2018-eccv-zero-shot} relies on the correspondences between dual-domain  data samples in   IRT to train the model. In contrast,  our method does not rely on such information thanks to the capability of our CoCoGAN in capturing the joint distribution of dual-domain images.

\section{Coupled Generative Adversarial Networks}
\label{sec:cogan}

The coupled generative adversarial networks (CoGAN) \cite{Liu-nips2016-CoupleGAN}  consists of two GANs, denoted as GAN$_1  $ and GAN$ _2 $, each of which corresponds to a domain. These two GANs have sharing  layers to deal with the high-level semantic concepts, and   individual layers to deal with low-level features for different domains.  This setting allows the two generators (or discriminators) to decode (or encode) the same high-level concepts by different ways in two domains.  

The CoGAN  captures the joint distribution of multi-domain images, and thus can  generate tuples of images, such as the RGB image and the depth image of the same scene.  
Different from the traditional methods that learn the joint distribution based on tuples of images, CoGAN is able to learn the joint distribution based on the images individually drawn from   marginal distributions. In other words, the training procedure does not rely on the correspondence between data samples in the two domains.

With GAN$_i  $ ($ i=1,2 $) consisting of generator g$ _i $ and discriminator $ f_i $, the training procedure of CoGAN optimizes the following minimax objective function
\begin{equation}
\label{eq:cogan-objective-function}
\begin{split}
\begin{aligned}
&\max\limits_{g_1,g_2}\min\limits_{f_1,f_2} V(f_1,f_2,g_1,g_2)  \equiv \\ &E_{x_1\sim p_{x_1}}[-\log f_1(x_1)]
+E_{z\sim p_z}[-\log (1-f_1(g_1(z)))]\\
+&E_{x_2\sim p_{x_2}}[-\log f_2(x_2)]
+E_{z\sim p_z}[-\log (1-f_2(g_2(z)))]
\end{aligned}
\end{split}
\end{equation}
subject to two constraints: 
\begin{itemize}
	\item 1) $ \theta_{g_1^j}=\theta_{g_2^j} $  \quad \quad \qquad  $ 1\leq j \leq s_{g} $
	\item 2) $ \theta_{f_1^{n_1-k}}=\theta_{f_2^{n_2-k}} $ \, \quad $ 0\leq k\leq s_{f}-1 $
\end{itemize}
where $ \theta_{g_i^j} $ denotes the parameter of the $ j $th layer in the generator $ g_i $ ($ i=1,2 $),   $ \theta_{f_i^{n_i-k}}$ denotes the parameter of the $ (k+1) $th layer from the last  in the  discriminator $ f_i $ ($ i=1,2 $), and $ n_i $ denotes the number of layers in the  discriminator $ f_i $. 
While the first constraint   indicates that the two generators have $ s_{g} $ sharing bottom layers, the second constraint  indicates that the two discriminators have $ s_{f} $ sharing top layers. With these two weight-sharing constraints, the two GANs can deal with high-level concepts in the same way, which is essential to learn the joint distribution  of data samples (i.e. $p_{x_1,x_2} $) based on the  samples drawn individually from the marginal distributions (i.e. $p_{x_1}$ and $p_{x_2}$) .

\begin{figure*}
	\centering
	\includegraphics[width=0.88\linewidth]{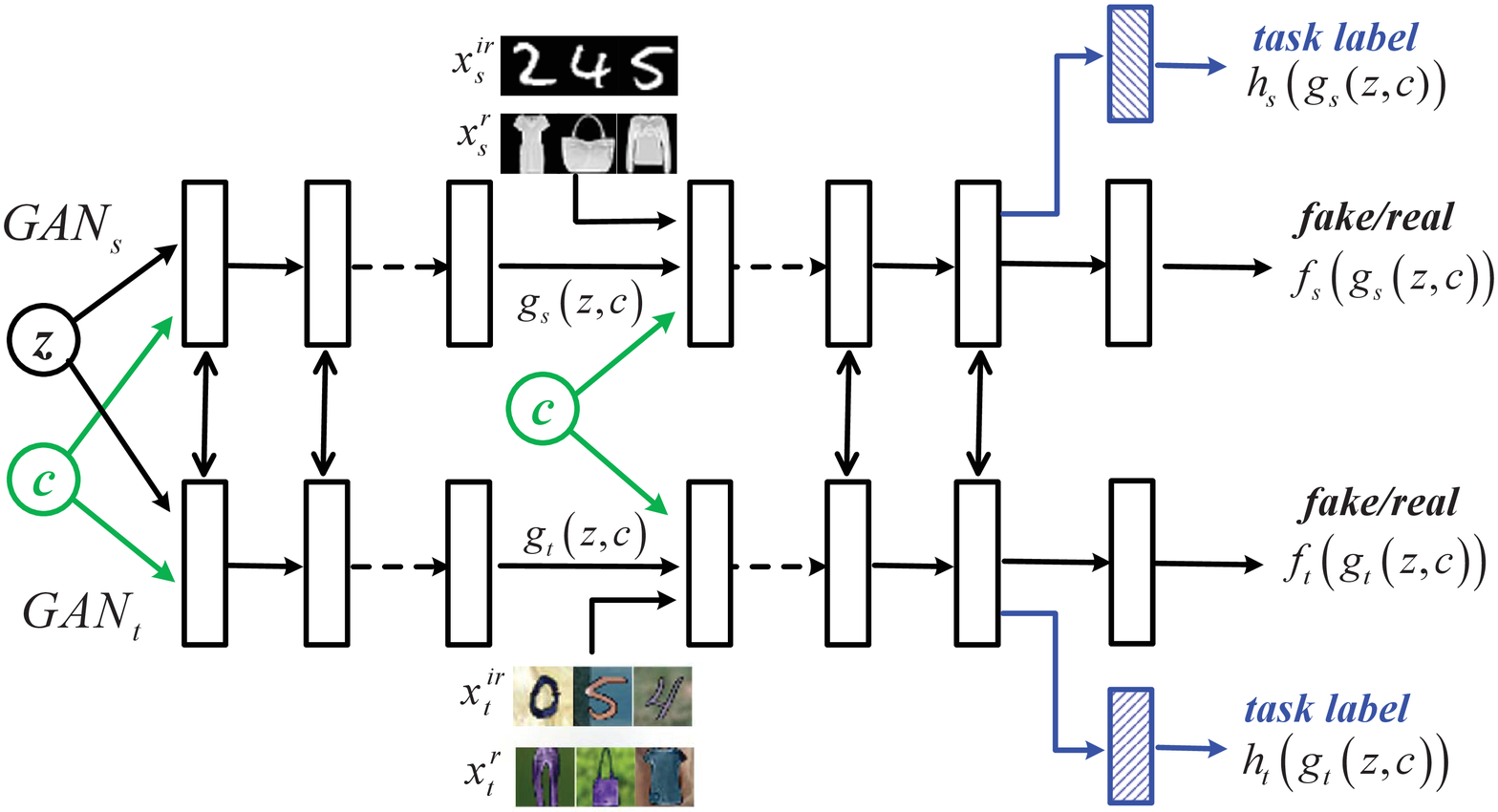}
	\caption{Illustration of the proposed CoCoGAN. The CoCoGAN extends CoGAN with a binary conditioning variable $ c $, which chooses RT/IRT for the network to deal with. The GAN$ _s $ process the source-domain data and the GAN$ _t $ process the target-domain data. The  double-headed arrows connect the sharing layers between the two branches. We maximize the loss of RT/IRT task label classifiers to obtain aligned representations across tasks.
	}  
	\label{fig:the-network-structureofCoCoGAN}
\end{figure*}

\section{Approach}
\label{sec:approach}

Motivated by the success of conditioning methods in various computer vision tasks \cite{Mirza2014ConditionalGA,Hong-inferring-cvpr2018,Wang2018pix2pixHD,PathakCVPR16context,Chrysos2018rocgan,Yang2018diversitysensitive},  we extend CoGAN into a conditioning model and propose conditional coupled generative adversarial networks (CoCoGAN).  
In order to train the CoCoGAN when the target-domain data for the task of interest are non-available and make it applicable to ZSDA tasks, we propose a new supervisory signal, i.e. the alignment between representations across tasks.

Our method involves two tasks, i.e. the relevant task (RT) and the irrelevant task (IRT). For each task, the data are from two domains, i.e. the source domain and the target domain. 
Let $ x_s^r $ and $ x_t^r $  be the samples from the source domain and the target domain for   RT, and let $ x_s^{ir} $ and $ x_t^{ir} $ be the samples for   IRT. We also use $ X_s^r $, $ X_t^r $, $ X_s^{ir} $ and $ X_t^{ir} $ to denote the sample sets, i.e. $ X_s^{r}=\{x_s^r\} $ etc. Given $ X_s^r $,  $ X_s^{ir} $ and $ X_t^{ir} $, the ZSDA task aims to learn a machine learning model for the non-available $ X_t^r $.

\subsection{CoCoGAN}

As shown in Fig. \ref{fig:the-network-structureofCoCoGAN}, our CoCoGAN extends the CoGAN to a conditional model by a binary conditioning variable $ c$, which chooses a task for the CoGAN to deal with. 
It deals with data samples in   IRT if $ c=0 $ , and  deals with data samples in   RT if $ c=1 $.
Our CoCoGAN uses a pair of GANs to capture the joint  distribution of data samples across two domains. 
Specifically, the GAN$_s  $  (GAN$_t  $)   processes source-domain (target-domain) samples with generator  $ g_s $ ($ g_t $) and discriminator $ f_s $ ($ f_t $).
The two generators $ g_s $ and $ g_t $ try to confuse the discriminators $ f_s $ and $ f_t $ by synthesizing pairs of samples that are similar to the real images as much as possible. In Fig. \ref{fig:the-network-structureofCoCoGAN}, we use double-headed arrows to connect the sharing layers, which allow us to learn the correspondences between dual-domain images.

When $ X_s^r $, $ X_t^r $, $ X_s^{ir} $ and $ X_t^{ir} $ are available, we can simply optimize the following objective function to train the CoCoGAN:
\begin{equation}
\label{eq:cocogan-objective-function}
\begin{split}
\begin{aligned}
&\max\limits_{g_s,g_t}\min\limits_{f_s,f_t} V(f_s,f_t,g_s,g_t)  \equiv \\ &E_{x_s\sim p_{x_s}}[-\log f_s(x_s,c)]
+E_{z\sim p_z}[-\log (1-f_s(g_s(z,c)))]\\
+&E_{x_t\sim p_{x_t}}[-\log f_t(x_t,c)]
+E_{z\sim p_z}[-\log (1-f_t(g_t(z,c)))]
\end{aligned}
\end{split}
\end{equation}
subject to two constraints: 
\begin{itemize}
	\item 1) $ \theta_{g_s^j}=\theta_{g_t^j} $  \quad \quad \qquad  $ 1\leq j \leq s_{g} $
	\item 2) $ \theta_{f_s^{n_1-k}}=\theta_{f_t^{n_2-k}} $ \, \quad $ 0\leq k\leq s_{f}-1 $
\end{itemize}
The source-domain sample $ x_s $ is drawn from the sample set $ X_s^{ir} $, if $ c=0 $; and drawn from  $ X_s^{r} $, if $ c=1 $.
Similarly,  the target-domain sample $ x_t $ is drawn from the sample set $ X_t^{ir} $, if $ c=0 $; and drawn from  $ X_t^{r} $, if $ c=1 $.
Given the data samples from two domains in the two tasks, we can easily train the CoCoGAN to capture the joint distribution of the dual-domain data samples.

\subsection{Representation Alignment}

In a ZSDA task, however, it is difficult to train the CoCoGAN  due to the non-availability of $ X_t^{r} $. If we simply optimize the objective function (\ref{eq:cocogan-objective-function}) with the available data, i.e. $ X_s^r $,  $ X_s^{ir} $ and $ X_t^{ir} $, the GAN$_t $ tends to be biased towards the IRT in the target domain and cannot well capture the distribution of the non-available target-domain data inside RT. To overcome such a problem, we propose an additional supervisory signal to train the CoCoGAN for ZSDA tasks, i.e.  the alignment of data sample representations across tasks.
In other words, we expect the representations from two different tasks are non-distinguishable from each other in a given domain.

Generally, the CoCoGAN aims to discover the correlation between the source domain and the target domain by capturing the joint distribution of the dual-domain samples for both IRT and RT. We can consider GAN$ _s $ and  GAN$ _t $  as conjugate in the two domains, as they are expected to generate sample pairs $ (x_s,x_t) $ with correspondence, i.e., $  x_t =T(x_s) $.  Here, $ T(.) $ is the transformation from  the source domain to the target domain.
In order to gain the ability to generate sample pairs $ (x_s,x_t)=(x_s,T(x_s)) $  with sharing high-level concepts (such as class label and semantic attributes), the processing procedure of GAN$ _s $ in the source domain and that of GAN$ _t $ in the target domain should have the same semantic meanings. 
Thus, the  representation extraction procedures, i.e  $ r_s(.) $ in $ g_s $ and  $ r_t(.) $ in  $ g_t $, should produce two representation sets with the same semantic meaning in a given task, which is denoted as   $ r_s(x_s,c) \simeq r_t(x_t,c) $ ($ c=0$ or $  1 $) in this paper. In other words, the representation of $ X_s^{ir} $ and that of $ X_t^{ir} $ share the semantic meanings in an ideal CoCoGAN, i.e.
\begin{equation}
\small
\label{eq:semantic-eqi-IRT}
r_s(X_s^{ir},c=0) \simeq r_t(T(X_s^{ir}),c=0) \equiv  r_t(X_t^{ir},c=0)
\end{equation}
Similarly, we also expect the representation of the non-available $ X_t^r $ share the semantic meanings with that of $ X_s^r $, i.e. 
\begin{equation}
\small
\label{eq:semantic-eqi-RT}
r_s(X_s^r,c=1) \simeq r_t(T(X_s^r),c=1) \equiv  r_t(X_t^r,c=1)
\end{equation}
Thus, if we explicitly align $ r_s(X_s^{ir},c=0)  $ and $ r_s(X_s^r,c=1) $ in the source domain, we can expect the alignment between $ r_t(X_t^{ir},c=0) $  and $  r_t(X_t^r,c=1) $ in the target domain. In other words, if  $ r_s(.) $ encodes samples for two different tasks with the same representation space in the source domain, then $ r_t(.) $ (i.e. the conjugation of $ r_s(.) $) should achieve the same goal in the target domain.

Based on the above analysis, we first  explicitly impose representation alignment across tasks in the source domain, and then take the representation alignment in the target domain as the supervisory signal to train the CoCoGAN. In this way, the generator $ g_t $ in GAN$ _t $ searches in the target domain to produce the samples whose representations are aligned with $ X_t^{ir} $ in the target domain.

\subsection{Training}

As shown in Fig. \ref{fig:the-network-structureofCoCoGAN}, we propose a binary RT/IRT task classifier for each tasks, i.e $ h_s(.) $ for the source domain and $ h_t(.) $ for the target domain,  to identify the involving task of the input. We  maximize the loss of these classifiers in order to achieve   representation alignment.  In other words, we expect that the representation of a sample in RT is indistinguishable from that of a  sample in   IRT if they belong to the same domain.  
Our objective functions for the tasks classifiers are given as follows
\begin{equation}
\small
\label{eq:loss-source-domain}
\max\limits_{h_s} L_s\equiv  E_{\substack{x_s\sim p_{x_s}}}[\ell(h_s(x_s))]+E_{\substack{z\sim p_{z}}}[\ell(h_s(g_s(z,c)))]
\end{equation}
\begin{equation}
\small
\label{eq:loss-target-domain}
\max\limits_{h_t}  L_t\equiv
E_{\substack{x_t\sim p_{x_t}}}[\ell(h_t(x_t))]+E_{\substack{z\sim p_{z}}}[\ell(h_t(g_t(z,c)))]
\end{equation}
The loss function $ \ell(.)  $ for the task classification (i.e. RT/IRT)  is the logistic function. 
Both of the two task classifiers are implemented with  convolutional neural networks.

In order to jointly optimize the Eq. (\ref{eq:cocogan-objective-function}), (\ref{eq:loss-source-domain}), and (\ref{eq:loss-target-domain}), we alternatively optimize the following two objective functions:
\begin{align}
\footnotesize
(\hat{f}_s,\hat{f}_t,\hat{h}_s,\hat{h}_t)=&\argmin \limits_{f_s,f_t,h_s,h_t} V(f_s,f_t,\hat{g}_s,\hat{g}_t)-(L_s+L_t)\label{eq:optimize-discriminator-classifier}
\\
(\hat{g}_s,\hat{g}_t)=&\argmax\limits_{g_s,g_t}  V(\hat{f}_s,\hat{f}_t,g_s,g_t) \label{eq:optimize-generator}
\end{align}
While Eq. (\ref{eq:optimize-discriminator-classifier}) updates the discriminators and the task classifiers with the fixed generators,
Eq. (\ref{eq:optimize-generator}) updates the generators with the fixed discriminators. With the updates in Eq. (\ref{eq:optimize-discriminator-classifier}), the representations are more discriminative in the real/fake classification task and less discriminative in the RT/IRT classification task. With the updates in Eq. (\ref{eq:optimize-generator}), the generators generate sample pairs which are more similar to the real data samples. We use the standard stochastic gradient method to optimize both Eq. (\ref{eq:optimize-discriminator-classifier}) and Eq. (\ref{eq:optimize-generator}).

\section{Experiments}
\label{sec:experiment}

\begin{figure*}[t]
	\begin{center}
		\includegraphics[width=0.9\linewidth]{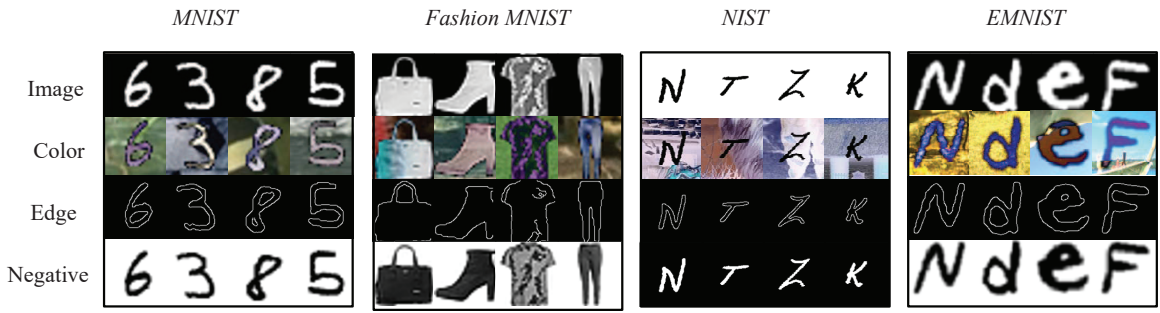}
	\end{center}
	\caption{The example images of  $ 4 $ datasets and their counterparts in  $ 3 $ different domains. The first row shows the original images. We use the method in \cite{Ganin-icml2015-unsupervised} to obtain the second row, a Canny detector to obtain the third row, and the negation procedure to obtain the fourth row.  }
	\label{fig:exampleimages}
\end{figure*}

\subsection{Datasets}
\label{subsec:dataset}
We evaluate our method on four  datasets, including MNIST \cite{mnist}, Fashion-MNIST \cite{fashion-mnist}, NIST \cite{nist}, and EMNIST \cite{emnist}.

The MNIST ($ D_M $) is proposed for  handwritten digit image analysis. This dataset has 60000 training and 10000 testing grayscale images. Every sample belongs to one of the $ 10 $ classes, i.e. from $ 0 $ to $ 9 $. The image size is $ 28 \times 28 $.

The Fashion-MNIST ($ D_F $)  is a dataset for fashion image analysis. It has the same size with MNIST, i.e.  60000 for training and 10000 for  testing. 
The image size is also $ 28 \times 28 $. The images  are  manually labeled by fashion experts  with one of the following $ 10 $ silhouette codes, i.e. \textit{T-shirt, trouser, pullover, dress, coat, sandals, shirt, sneaker, bag}, and \textit{ankle boot}.

NIST ($ D_N $) is a handwritten letter image dataset. In our experiment, we use the images of both uppercase and lowercase letters. In total, we have  $ 387361 $ training and $ 23941 $ testing images  from  $ 52 $ different classes. The image size is $ 128 \times 128 $. This dataset is imbalanced and there are large differences in the occurrence frequencies for the $ 52 $ classes.

EMNIST ($ D_E $) is an extension of NIST. To be different from the NIST, we merge the uppercase and lowercase letters to form a balanced $ 26 $-class   dataset. This subset has $ 124800 $ training and $ 20800 $ testing images. The image size is $ 28 \times 28 $.

All these four different datasets consist of gray-scale images, and we consider them in the  gray domain ($ G\textendash domain $). In order to evaluate our method, we create    three more domains via transformations, i.e. the colored domain ($ C\textendash domain $), the edge domain ($ E\textendash domain $), and the negative domain ($ N\textendash domain $).  Firstly, we transform the gray-scale images into color images using  the method proposed in \cite{Ganin-icml2015-unsupervised}. 
For a given image $ I \in R^{m\times n} $, we  randomly crop a patch  $ P \in R^{m\times n} $ from a color image in BSDS500 \cite{Arbelaez:2011:CDH:1963053.1963088}, and combine them together by $ I_c=|I-P| $ in each channel. Secondly, we transform a gray-scale image $ I $ into an edge image  $ I_e $ with a canny detector. 
Thirdly, we obtain the negative of each image by    $ I_n=255-I $.
Fig. \ref{fig:exampleimages} shows example images in four different domains.

\newcommand{\tabincell}[2]{\begin{tabular}{@{}#1@{}}#2\end{tabular}}

\begin{table*}[tb]
	\centering
	\caption{The classification accuracies of the proposed method and the baselines with $ 5 $ different settings of (source domain, target domain) pairs. We remove the task classifiers in both source domain and target domain  from the \textit{CoCoGAN}  to create the baseline  \textit{CoCoGAN w/o T}}
	\begin{tabular}{l|p{0.8cm}p{0.8cm}p{0.8cm}|p{0.8cm}p{0.8cm}p{0.8cm}|p{0.8cm}p{0.8cm}|p{0.8cm}p{0.8cm} }
		\multicolumn{11}{c}{A. (source domain, target domain)$= (G\textendash domain, C\textendash domain) $} \\ 
		\hline 
		RT & \multicolumn{3}{c|}{MNIST ( $ D_M $)} & \multicolumn{3}{|c|}{Fashion-MNIST ($  D_F$) } & \multicolumn{2}{c|}{NIST ($ D_N $)} & \multicolumn{2}{|c}{EMNIST ($ D_E $)} 
		\\ 		
		\cline{2-11}
		IRT &  $  D_F$  & $ D_N $ & $ D_E $ & $ D_M $ &  $ D_N $ & $ D_E $ & $ D_M $ & $  D_F$   & $ D_M $ & $  D_F$ 
		\\ 	\hline \hline
		ZDDA & 73.2  & 92.0  & 94.8  &  51.6 & 43.9 & 65.3 &  34.3 & 21.9 &  71.2 &  47.0
		\\ 
		CoCoGAN w/o T  &  68.3  &  81.6 &  74.7 &  39.7 & 48.2 & 55.8  &  35.2  & 38.8 & 46.7  &   41.8
		\\ 
		CoCoGAN & 78.1  &  92.4 &  95.6   &  56.8 & 56.7  & 66.8 &  41.0 &  44.9&  75.0 &  54.8  
		\\ 
		\hline
		\multicolumn{11}{c}{ } \\  
		\multicolumn{11}{c}{B. (source domain, target domain)$= (G\textendash domain, E\textendash domain) $} 
		\\ 		
		\hline 
		RT & \multicolumn{3}{c|}{MNIST ( $ D_M $)} & \multicolumn{3}{|c|}{Fashion-MNIST ($  D_F$) } & \multicolumn{2}{c|}{NIST ($ D_N $)} & \multicolumn{2}{|c}{EMNIST ($ D_E $)} 
		\\ 		
		\cline{2-11}
		IRT &  $  D_F$  & $ D_N $ & $ D_E $ & $ D_M $ &  $ D_N $ & $ D_E $ & $ D_M $ & $  D_F$   & $ D_M $ & $  D_F$ 
		\\ 	\hline \hline
		ZDDA & 72.5  &  91.5  &  93.2 &  54.1  & 54.0  &  65.8 & 42.3  &  28.4  &  73.6  &   50.7   
		\\ 
		CoCoGAN w/o T &  67.1  &  74.8 &  81.5 &  47.5 &  50.2 & 56.1 &  41.2 &  30.9 &  63.6 &   51.9
		\\ 
		CoCoGAN & 79.6  & 94.9  &  95.4   &   61.5 & 57.5 &   71.0 & 48.0 & 36.3 &  77.9 &   58.6
		\\ 
		\hline
		\multicolumn{11}{c}{ } \\  
		\multicolumn{11}{c}{C. (source domain, target domain)$= (G\textendash domain, N\textendash domain) $} 
		\\
		\hline 
		RT & \multicolumn{3}{c|}{MNIST ( $ D_M $)} & \multicolumn{3}{|c|}{Fashion-MNIST ($  D_F$) } & \multicolumn{2}{c|}{NIST ($ D_N $)} & \multicolumn{2}{|c}{EMNIST ($ D_E $)} 
		\\ 		
		\cline{2-11}
		IRT &  $  D_F$  & $ D_N $ & $ D_E $ & $ D_M $ &  $ D_N $ & $ D_E $ & $ D_M $ & $  D_F$   & $ D_M $ & $  D_F$ 
		\\ 	\hline \hline
		ZDDA & 77.9  & 82.4  &  90.5 &  61.4  &  47.4  &  62.7  &  37.8  &  38.7  &  76.2  &    53.4  
		\\ 
		CoCoGAN w/o T  & 62.7 & 67.3  &  72.8 &  51.8  & 47.5 & 51.2 &  39.3 & 36.7 & 60.8 &  39.1
		\\ 
		CoCoGAN & 80.3  &  87.5 &   93.1  & 66.0 &  52.2 &  69.3 &  45.7 &  53.8&  81.1 &  56.5 
		\\ 
		\hline
		\multicolumn{11}{c}{} \\  
		\multicolumn{11}{c}{D. (source domain, target domain)$= (C\textendash domain, G\textendash domain) $}	
		\\ 		
		\hline 
		RT & \multicolumn{3}{c|}{MNIST ( $ D_M $)} & \multicolumn{3}{|c|}{Fashion-MNIST ($  D_F$) } & \multicolumn{2}{c|}{NIST ($ D_N $)} & \multicolumn{2}{|c}{EMNIST ($ D_E $)} 
		\\ 		
		\cline{2-11}
		IRT &  $  D_F$  & $ D_N $ & $ D_E $ & $ D_M $ &  $ D_N $ & $ D_E $ & $ D_M $ & $  D_F$   & $ D_M $ & $  D_F$ 
		\\ 	\hline \hline
		ZDDA  & 67.4  &  85.7 &  87.6  &  55.1  &  49.2  &  59.5 &  39.6  &  23.7  &  75.5  &   52.0   
		\\ 
		CoCoGAN w/o T &  54.7 &  69.0  &  63.5 &  43.4  & 40.6 & 51.6 &  21.4  & 30.9 & 49.5 & 48.2 
		\\ 
		CoCoGAN & 73.2 & 89.6  &  94.7 &  61.1 & 50.7 & 70.2& 47.5  & 57.7 & 80.2&  67.4
		\\ 
		\hline
		\multicolumn{11}{c}{} \\  
		\multicolumn{11}{c}{E. (source domain, target domain)$= (N\textendash domain, G\textendash domain) $}	
		\\ 		
		\hline 
		RT & \multicolumn{3}{c|}{MNIST ( $ D_M $)} & \multicolumn{3}{|c|}{Fashion-MNIST ($  D_F$) } & \multicolumn{2}{c|}{NIST ($ D_N $)} & \multicolumn{2}{|c}{EMNIST ($ D_E $)} 
		\\ 		
		\cline{2-11}
		IRT &  $  D_F$  & $ D_N $ & $ D_E $ & $ D_M $ &  $ D_N $ & $ D_E $ & $ D_M $ & $  D_F$   & $ D_M $ & $  D_F$ 
		\\ 	\hline \hline
		ZDDA  & 78.5 & 90.7  & 87.6  &  56.6 & 57.1 & 67.1 & 34.1 & 39.5 & 67.7 &  45.5     
		\\ 
		CoCoGAN w/o T & 66.1 &  75.9 &  76.3 &  49.9 & 53.1 & 58.7 &  35.6 & 33.7 & 53.0 &  32.5
		\\ 
		CoCoGAN & 80.1  & 92.8  &  93.6  & 63.4 &  61.0 & 72.8 & 47.0 & 43.9 & 78.8  & 58.4 
		\\ 
		\hline
	\end{tabular}
	\label{tab:results}
\end{table*}

\subsection{Implementation details}

We denote our method as \textit{CoCoGAN}, and compare   it with two baselines. The first baseline is \textit{ZDDA} \cite{Peng-2018-eccv-zero-shot}, which is the only work that adopts deep learning technique for ZSDA. To verify the effectiveness of the alignment between representations as the supervisory signal, we take   the CoCoGAN without any task classifier as the second baseline, and denote it as \textit{CoCoGAN w/o T }in   this work.

Our CoCoGAN is implemented with convolutional neural networks and  its two branches (i.e. GAN$ _s $ and GAN$ _t $) have the same network structure. 
The generator has  $ 7 $  transposed convolutional layers to decode the random vector $ z $ into a realistic sample for   RT if $ c=1 $ and for  IRT if $ c=0 $. For   representation learning from the real  and generated images, the discriminators have $ 5 $ convolutional layers with stride 2,  which are denoted as $ r_s(\cdot) $ in the source domain and $ r_t(\cdot) $ in the target domain. In addition, the discriminators  have two convolutional layers for fake/real classification. 
Thus, both the generators and the discriminators have $ 7 $ layers. 
The binary  classifiers, i.e. $ h_s(\cdot) $ and  $ h_t(\cdot) $, use two fully connected layers to classify $ r_s(x_s) $ and  $ r_t(x_t) $ into RT or IRT.

In the training stage, we partition the sample set  of IRT into two non-overlapping halves in each domain, i.e. $ X_s^{ir}=X_{s1}^{ir}\cup X_{s2}^{ir} $ and $ X_t^{ir}=X_{t1}^{ir}\cup X_{t2}^{ir} $, where $ X_{s1}^{ir} \cap X_{s2}^{ir} =\varnothing $, $ X_{t1}^{ir}\cap X_{t2}^{ir} =\varnothing$, $ X_{t1}^{ir} =T(X_{s1}^{ir})   $, and $ X_{t2}^{ir} =T(X_{s2}^{ir}) $. We use the   first half in the source domain (i.e. $X_{s1}^{ir}  $)  to train  GAN$ _s $ and use the second half  in the target domain (i.e. $X_{t2}^{ir}  $)  to train GAN$ _t $. Thus, there is no correspondence between the source-domain samples and the target-domain samples.
Compared with our proposed, the ZDDA \cite{Peng-2018-eccv-zero-shot} instead needs the correspondence between data samples in the training procedure. We use the sample set with correspondence to train ZDDA, i.e. $ X_{s1}^{ir} \cup X_{t1}^{ir} $  or $ X_{s2}^{ir} \cup X_{t2}^{ir} $.


We use the trained CoCoGAN for image classification in the target domain   of   RT and obtain the classifier using the following three steps. 
Firstly, we generate a set of sample pairs with correspondence  $ (\tilde{x}_s^r,\tilde{x}_t^r)  $ using the source-domain generators $ g_s $ and the target-domain generator $ g_t $.   Secondly, we train a label predictor $ C_s(x_s^r) $ for the source-domain samples in   RT based on  the available $ X_s^r $ and their labels, and use this predictor to obtain the sharing labels of the generated samples, i.e. $ label_{\tilde{x}_t^r}  =label_{\tilde{x}_s^r}=C_s(\tilde{x}_s^r)$. Thirdly, we train a label classifier for the target-domain samples in  RT based on the generated samples and their labels.

\subsection{Results}

In order to evaluate the proposed CoCoGAN, we have five different pairs of source domain and target domain. On one hand, we take  $ G\textendash domain $ as the source domain and take the other three domains as the target domain. Thus, the source and target domain pairs are $ (G\textendash domain, C\textendash domain) $, $ (G\textendash domain, E\textendash domain) $,  and $ (G\textendash domain, N\textendash domain) $.   On the other hand, we also take  $ G\textendash domain $ as the target domain and transfer knowledge from the other two domains, where the dual-domain pairs are  $(C\textendash domain, G\textendash domain) $ and $ (N\textendash domain, G\textendash domain) $.

The four datasets in Sec. \ref{subsec:dataset} involves  three different tasks, i.e. digit image classification, fashion image classification, and letter image classification.  Given the RT, we can take any of the other two as the IRT. The   NIST and the EMNIST share the task since both of them consist of letter images. Thus, we do not  take \textit{(NIST, EMNIST)} or \textit{(EMNIST, NIST)} as valid \textit{(RT,IRT)} pair  in our experiments.

Tab. \ref{tab:results} lists the classification accuracies of different settings. 
As seen, our method performs significantly better that ZDDA \cite{Peng-2018-eccv-zero-shot}.  
Taking NIST classification in Tab. (\ref{tab:results})-D as an example, our proposed CoCoGAN outperforms ZDDA by $ 7.9\% $ when  the IRT is digit image analysis and by $ 34.0\% $  when the IRT is fashion image analysis. 
The comparative results demonstrate that our proposed indeed obtain discriminative representations from the target-domain data of RT based on the representation extraction procedure learned in CoCoGAN.  In addition, our method has more potential applications than ZDDA, whose performance is heavily relied on the correspondence between dual-domain samples in the IRT.

Our proposed CoCoGAN beats the baseline \textit{CoCoGAN w/o T} by $ 15.6\% $ on average, indicating the effectiveness of the task label classifiers in adapting the GAN$_t  $ towards the RT.  Without the task label classifiers, the non-sharing layers in both generator $ g_t $ and  discriminator $ f_t $  are trained solely by the samples in IRT, and thus not suitable for the non-available target-domain data in RT. In order to make them applicable to the target-domain data in RT, our CoCoGAN updates the parameter of  these non-sharing layers  based on the correlation between the two domains, i.e. the representation alignment across tasks in this work. It is these supervisory signals that guide the generators to decode and the discriminator to encode the low-level features of those non-available samples properly.

Our method also beats many existing methods which rely on the availability of the target-domain data samples in the training procedure. Taking $ C \textendash domain $ as the source domain and  $ G \textendash domain $ as the target domain, our method achieves the accuracy of $ 94.7\% $  on MNIST, yet the accuracies of the existing techniques are:  $ 86.7\% $ in  \cite{Sener-nips-learning-transferrable},  $ 89.5\% $ in   \cite{Haeusser-ICCV2017-associative}, and  $ 94.2\% $ in \cite{SaitoUH17-icml2017}, respectively.

\begin{figure}[tb]
	\begin{center}
		\includegraphics[width=0.99\linewidth]{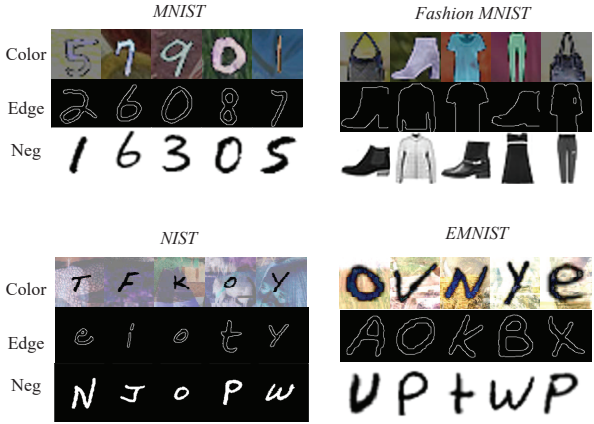}
	\end{center}
	\caption{The generated non-available target-domain images by CoCoGAN in the $ C\textendash domain $,  $ E\textendash domain $, and $ N\textendash domain $. These images are in the same style with the real images.}
	\label{fig:generatedImages}
\end{figure}

\begin{table}[htb]
	\caption{Taking $ G\textendash domain $  as the source domain, the average overlap ratios between the generated targe-domain images and the ones obtained by the procedure described in Sec. \ref{subsec:dataset} }
	\begin{tabular}{lcccc}	 
		\multicolumn{5}{c}{A. The overlap ratios in $ E\textendash domain $}         \\
		\hline
		& $ D_M $ & $ D_F $ & $ D_N $ & $ D_E $ \\
		CoCoGAN w/o T & 0.816  & 0.707  &  0.727 &  0.749 \\
		CoCoGAN       & 0.873 & 0.786 & 0.803 & 0.812    \\
		\hline
		\\
		\multicolumn{5}{c}{B. The overlap ratios in $ N\textendash domain $}         \\
		\hline
		& $ D_M $ & $ D_F $ & $ D_N $ & $ D_E $ \\
		CoCoGAN w/o T &  0.804 & 0.772  & 0.704  &  0.733 \\
		CoCoGAN       &  0.863 & 0.824 & 0.844 & 0.812 
		\\
		\hline
	\end{tabular}
	\label{tab:overlap}
\end{table}
In order to show the capability of the proposed CoCoGAN in capturing the joint distribution of dual-domain images, we visualize some generated samples in
Fig. \ref{fig:generatedImages}.  We also use the method proposed in \cite{Liu-nips2016-CoupleGAN} to evaluate the correspondence between the generated sample pairs with three steps. The first step generates a set of sample pairs $ (\tilde{x}_s^r,\tilde{x}_t^r) $ based on the trained CoCoGAN by changing the random variable $ z $. The second step produces the target-domain correspondence, i.e.  $ T(\tilde{x}_s^r) $, for  the  source-domain sample  $ \tilde{x}_s^r $  by using  the method described in Sec. \ref{subsec:dataset}. The third step calculates the overlap ratio between $ \tilde{x}_t^r $ and  $ T(\tilde{x}_s^r) $. Taking the gray-scale images as the source domain, Tab. \ref{tab:overlap} lists the average overlap ratios in the edge domain and the negative domain. The higher the overlap ratio, the more accurate the correspondence between the generated sample pairs. As generating a color image involves a random patch sampling process, this metric becomes meaningless in color domain.  As seen in the Tab \ref{tab:overlap}, our proposed CoCoGAN achieves higher overlap ratios than the baseline \textit{CoCoGAN w/o T}, indicating the proposed supervisory signal improves the correspondence between the dual-domain samples.

\section{Conclusion}
\label{sec:conclusion}

Zero-shot domain adaptation refers to the problem where the target-domain data are not available in the training stage.  We  propose a so-called CoCoGAN to solve this problem by extending the CoGAN into a conditioning model.  Essentially, our CoCoGAN consists of two GANs in order to capture the joint distribution of data samples across two domains and two tasks. 
The model for the unseen target-domain data in RT is learned based on the source-domain data in RT and the dual-domain data in an IRT. 
While the former provide the high-level concepts of the unseen target-domain data, the latter carry the sharing correlation between the two domains  in RT and IRT.
To train the CoCoGAN in the absence of the target-domain data, we introduce a new supervisory signal, i.e. the alignment between representations across tasks. 
In comparison with the existing methods such as \cite{Peng-2018-eccv-zero-shot}, our method does not rely on the correspondences between samples in IRT, and thus has more potential applications.
Extensive experiments are carried out on four publicly available datasets, and the results validate  the effectiveness of our proposed method in generating the non-available data samples and extracting their representations.

\section*{Acknowledgment}
The authors wish to acknowledge the financial support from: (i) Natural Science Foundation China (NSFC) under the Grant No. 61620106008 and No. 61802266; and (ii) Shenzhen Commission for Scientific Research \& Innovations under the Grant No. JCYJ20160226191842793.

{\small
	\bibliographystyle{ieee_fullname}
	\bibliography{cocoganbib}
}

\end{document}